\RequirePackage{fix-cm}
\documentclass[smallextended]{svjour3}       
\smartqed  
\usepackage{graphicx}
\usepackage{url}

\begin{document}

\title{Conditions for Normative Decision Making at the Fire Ground} 
\subtitle{PREPRINT * Short Communication}

\author{Adam Krasuski}

\institute{A. Krasuski \at
              The Main School of Fire Service \\
			  Slowackiego 52/54\\
              01-629 Warsaw, Poland\\
              \email{krasuski@inf.sgsp.edu.pl}           
}

\date{Submitted: 15.12.2015}
\maketitle
\begin{abstract}

We discuss the changes in an attitude to decision making at the fire
ground. The changes are driven by the recent technological shift. The
emerging new approaches in sensing and data processing (under common
umbrella of Cyber-Physical Systems) allow for leveling off the gap,
between humans and machines, in perception of the fire ground.
Furthermore, results from descriptive decision theory question the
rationality of human choices. This creates the need for searching and
testing new approaches for decision making during emergency. We propose
the framework that addresses this need. The primary feature of the
framework are possibilities for incorporation of normative and
prescriptive approaches to decision making. The framework also allows
for comparison of the performance of decisions, between human and
machine. 

\keywords{Fire Service \and Decision Support \and Decision Theory \and Sensory Data}
\end{abstract}
\section{Introduction}

\emph{In this Section we outline the main obstacles for machines to
replace the incident commander.} 

The fire ground management attracts many researchers from different
domains. This very complex decision making environment is often
considered as the \emph{iron wall}\footnote{The nickname of Tigran
Petrosian recognized as the hardest chess player to beat ever.} for many
approaches from decision theory and artificial intelligence. The vaguely
defined goals, high uncertainty, dynamically changed conditions, time
and mental pressure as well as team and resources constrains, cause that
most of the approaches to this field, fail to be successful. A
well-trained human is still unrivaled decision-maker in this
environment.

The reasons underlaying the superiority of humans are: (1) Exceptional
perception -- humans with their senses, and the ability of dynamically
changing focus, information fusion, can still extract and infer more
information than any set of sensors. (2) Incredible pattern recognition
abilities -- humans have the ability to find patterns and cues in a very
noisy environment. (3) Dominant general knowledge about the world,
allowing for deduction and implementation of solutions from other
domains. (4) Bundle of heuristics for dealing with the complexity and
ambiguity. Many machine's algorithms, even accurate, fail to be used due
to the huge computational requirements\footnote{compare i.e. FDS
simulations}. The human is capable to dynamically structure and divide
the problem and solve it efficiently. (5) Legal and ethical constrains.
Humans may use solutions to some problems (i.e. triage) which are
resolved by ethics and philosophy~\cite{sorell2003ii}. 

All these issues restrict the ability for support (or replacement) the
incident commander by the machine. While rational choices can be made
based on normative and prescriptive approaches~\cite{keller1989role},
the application of these methods is limited by the complexity of the
environment and human processing capabilities. This led to the excessive
exploit of the naturalistic decision making~\cite{klein2008naturalistic}
at the fire ground. 

However, the recent pervasive application of sensors, global integration
of systems~\cite{atzori2010internet,khaitan2014design} as well as new
approaches for data processing\footnote{see i.e. big data or deep
learning} extended the perception and processing abilities of the
machines. Moreover, psychological works in the descriptive decision theory field,
revealed many pitfalls\footnote{The exhaustive list of cognitive biases
can be found at:
\url{https://en.wikipedia.org/wiki/List_of_cognitive_biases}} in human
decision making. All these results create possibilities and needs for
searching other than naturalistic decision making solutions, eligible
also to be executed on machines. 

In this article we propose the framework for searching for new solutions
for decision making at the fire ground. The framework incorporates
last achievements in sensing as well as artificial intelligence. The
data-flow within this framework allows for introduction and
investigation of normative and prescriptive decision theory approaches.
The framework defines a new promising direction of research, not
feasible previously.  

\section{Ill-informed Machines}\label{info}

\emph{In this Section we argue that currently the machines are capable of
enough perception to trigger further changes in decision making.} 

The machines' low perception combined with inability to dynamically
change focusing on the crucial things, is perceived as the most
challenging issue in overcoming the dominance of the humans at the
fire ground. These limitations established the opinion that the most
promising direction to improve performance of the management, is
increasing the situational awareness of the incident
commander~\cite{grorud2008national,blandford2004situation,gorman2006measuring}.
There are several works which investigate this
aspect~\cite{mendoncca2009group,cowlard2010sensor,carroll2006awareness,krasuski2013sensory}.
There are also few projects which tried to implement these
approaches~\cite{ashish2008situational,krasuski2014framework,sff}. 

ICRA\footnote{\url{http://www.icra-project.org/}} is one of such projects
where the focus is set to increase the situational awareness of the
incident commander. The general method to achieve the goal was building
the excessive data/sensory layer combined  with algorithms that convert
the low-level data (i.e. temperature distribution in the room) into
high-level concepts, easy to comprehend by humans (i.e. respiratory
risk). 

The ICRA data layer allows for gathering information related to the
following issues: (1) Full information about the building, organized
into Building Information
Modelling\footnote{\url{https://en.wikipedia.org/wiki/Building_information_modeling}}
(BIM). (2) The physical parameters of the environment inside the
building extracted from the building infrastructure in a form of: CO
concentration, temperature, optical density of air (measured in UV, IR).
(3) Parameters of the environment measured by deployed mobile robots in
a form of: $CO$, $CO_2$, $O_2$, $C_xH_y$ concentration, temperature and
visioning by IR and VIS cameras. (4) Indoor localization of the
civilians by usage of their mobile phone tracking
system~\cite{korbel2013locfusion}. (5) Firefighters indoor localization
based on dead-reckoning system~\cite{meina2015}. (6) Firefighters
activity reporting system based on the body sensors
network~\cite{meina2015tagging}. (7) Psychophysical conditions of the
firefighters measured by wearable sensors in a form of: ECG, breathing
rate, skin temperature and
conductance\footnote{\url{http://www.equivital.co.uk/about}}. (8)
Parameters of the equipments used by firefighters, including their
deployment time and disturbance factors (i.e. drop of pressure in fire hoses
depending on how they were deployed). (9) Database of Incident Data
Reporting System which allows among other for calculation of probability
distribution for various events. (10) Domain ontologies and what-if
database. The ontology structures the general knowledge about the domain
and allows for simple deductive inference about the facts. What-if is a
collection of experienced commander reflections about the outcomes
of the given decision against the states of nature.

The full availability of this data needs some pre-processing and depends
on proper robots deployment. However there is no limitation regarding
the buildings instances. 

Based on large collection of data, the fundamental questions
arise: \emph{Is it humans that can best utilise this data and produce
high-performance decisions? Shall we limit our activities to properly
present this information to incident commander or should we try
to suggest decisions?}

\section{Heuristics and Biases}

\emph{In this Section we question that human is the best decision-maker at
the fire ground.}

The current training programs for incident commanders are mostly
influenced\footnote{see for example FEMA
courses~\url{https://www.firstrespondertraining.gov/ntecatalog}} by the
naturalistic decision making (NDM)
approach~\cite{klein2008naturalistic}. The central concept in NDM is
\emph{expert intuition}. The NDM researchers try to learn from expert
professionals by identifying the cues and methods that experts use to
make their judgments. It originates from the observation that chess
grand masters have unusual ability to analyse complex positions and
quickly judge a line of play~\cite{chase1973mind}. 

The same performance was discovered in incident commanders required to
make decisions under conditions of uncertainty and time
pressure~\cite{klein1986rapid}. The recognition-primed decision (RPD)
was proposed as an effective strategy for decision making at the
fire ground~\cite{klein1986rapid}. The RPD and other NDM models
offer a generally encouraging picture of expert performance in multiple
domains~\cite{klein1999sources}.

The other approach -- heuristic and biases (HB) -- is in sharp contrast
to NDM. HB favours a skeptical attitude toward expertise and expert
judgments. The origins of this attitude can be traced to a Meehl's
monograph~\cite{meehl149clinical}, extended later by Kahneman and
Tversky~\cite{tversky1974judgment}. They described the simplifying
shortcuts of intuitive thinking and explained some 20 biases as
manifestations of these heuristics. 

In 2009 Kahneman and Klein -- top researchers from the two opposite
domains (NDM and HB) -- did a joint analysis for settling an agreement
in an intuitive expertise~\cite{kahneman2009conditions}. The researchers
agree that intuitive judgments can arise from genuine skill. However,
they can also arise from inappropriate application of the heuristic
processes. An environment of high validity is a necessary condition for
the development of skilled intuitions. The skilled intuitions are very
hard to achieve, need tens of thousand hours with high-quality
feedback. As a comparison, the top experienced commander form Warsaw
city has 1256 hours of commanding in total, with mid-quality feedback as
presented in~\cite{krasuski2012method}. Skilled intuitions are also very
vulnerable to fragmented expertise i.e. development of the  skills only
for narrowed type of fires. Moreover, the Kahneman and Klein suggest
that algorithms significantly outperform humans when validity of the
environment is low, or in highly predictable environments. 

It indicates that supremacy of human as a decision-maker is questionable
and needs further investigations. While humans mostly outperform
machines in the evaluation of the fire ground, this is only
part of the commanding process. However, \emph{budgeting}, i.e. proper
calculation of resources, deployments methods and time, calculation of
amount of needed water and many others are just as important. In these
calculations machines are faster and more accurate. 

\section{Normative Approach to Decision Making}

\emph{In this Section we propose a framework that takes advantage of
sensory data, models of phenomena, artificial intelligence and power of
rational choices ensured by normative decision theory.} 

The data-gathering ability of ICRA presented in Section~\ref{info}
allows for estimation of many subsequent parameters. Fire dynamics may
be calculated by using the models such
as~\cite{jahn2012forecasting2,overholt2012characterizing,fliszkiewicz2014evaluation}.
Output from these models, BIM data, localization of the civilians and
the evacuation models allow for estimation of the possible course of
action or: \emph{states of nature} in other notation. Moreover, the data
from Incident Data Reporting System allows for evaluation of
distributions of \emph{probabilities} of these states. 

The data from firefighters localization, activity reporting, and
psychophysical conditions, allow for better judgement of the real
human-resource availability. When combined with the parameters of the equipments 
it allows for calculating the sets of \emph{decision packages} available to the
incident commander. 

Ontology and what-if analysis knowledge, enables the calculation of the
\emph{outcomes} for subsequent decision packages with respect to the
states of nature. 

The outlined concepts (states of nature, decision packages, outcomes and
sparsely probabilities) are central for the normative and prescriptive
decision theory. The concepts establish the Decision Matrix (DM). DM
consists of: (a) rows related to the possible state of nature, (b)
columns represent decision available for decision-maker, (c) outcomes --
intersections of rows and columns -- represent the results for
combination of the decisions and the states of nature. The
distributions, if available, allow for calculating the possibilities of
occurrence of the states of nature. 

\begin{figure}[h] 
\begin{center}
\includegraphics[width=0.5\textwidth]{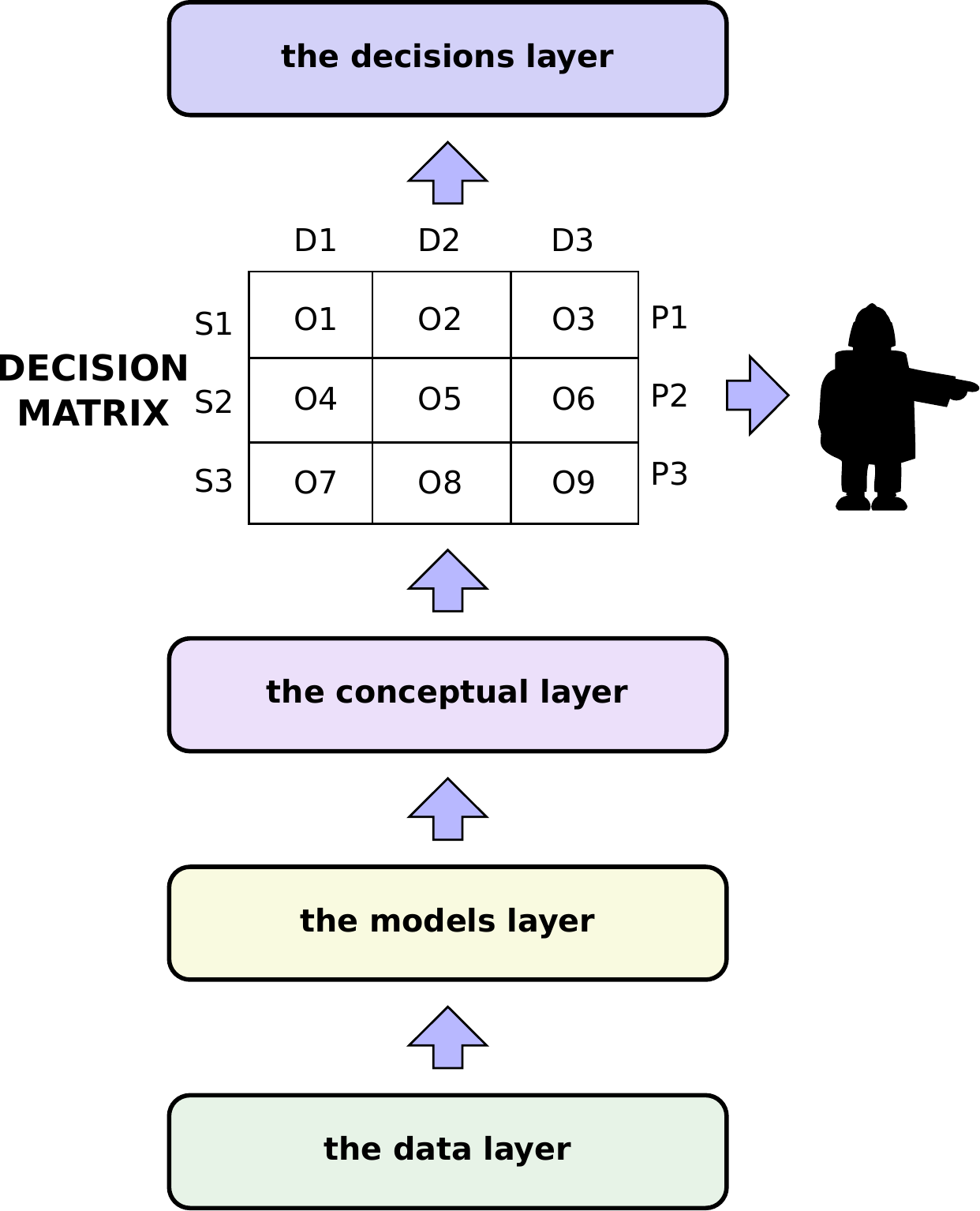} 
\end{center} 
\caption{The framework for decision support at the fire ground.
S1...S3 -- states of nature, D1...D3 decisions, O1...O9 outcomes,
P1...P3 probabilities of the consecutive states of nature.}
\label{scene} \end{figure}

Such a formalization  allows for introduction of normative and
prescriptive decision theories to the decision process.  The normative
approach allows for drawing a rational choice to the problem, eligible
at the emergency. The general framework for supporting (or even
eliminating) incident commander with normative approach is presented on
Figure~\ref{scene}. The framework also enables the comparison of
performance in decision making between the human and the machine. 

\emph{The data layer} represents the data resources which feed the
framework. It could be at least the data outlined in the
Section~\ref{info}. The layer represents features or variables which
are constant over time (ontologies, what-if etc.) and features
that are measured by sensors. 

\emph{The models layer} contains models and algorithms for calculation
of both physical and abstract variables. This layer employs the models for
fire modelling, evacuation models, finite elements methods, classifiers
for localization of the firefighters or any appropriate models from
engineering, physics and thermodynamics. The elements of the models
layer need an input, mostly derived from the data layer.  

\emph{The conceptual layer} consists mostly in models derived from
artificial intelligence (classification, clustering, granular computing,
adaptive judgements). The goal of the layer is to find the
characteristic patterns in the output from both models and data layer.
The patterns are used for approximation of the concepts related to the
states of nature, the possible decisions and the outcomes. 

The Decision Matrix is a central point of the framework. It separates
the process of information gathering from the decision making one. It
can be used for applying the normative decision making as well as
for presenting the gathered information to the incident commander.

\emph{The decisions layer} is a set of methods, rules and criteria from
normative and prescriptive domains, which allows for making the decision
according the rational choice rule.  

\section{Conclusion}

The emerged paradigms of Cyber-Physical Systems enable for gathering and
processing large amount of data to the extent not seen before. This
significantly increases the perception of the machines. The approaches
face the obstacles that keep the status-quo (naturalistic decision
making) of the decision making during emergency. On the other side, the
results from the descriptive decision theory contradict the appraisal
that human mostly makes rational choices. This enforced for searching
new solutions based on the computer systems and rational choices. A
normative decision theory is strongly rooted in the rational choice
paradigm. It advises how people should really make decisions.  

The paper proposes a framework for searching new solutions in the
normative and proscriptive domains. It allows also for the comparison of
human (naturalistic) decision making and the computer-aided one
(normative). The foundation for this framework derives from the ICRA
project. In the project the decision matrix was used to present the
information to the incident commander~\cite{krasuski2014framework}. This
initially proves the foundation of the proposed idea and opens the
possibilities for searching a better solution for decision making. The
outcomes from these researches may even allow for the elimination of the
human from decision making process.  



\begin{thebibliography}{10}
\providecommand{\url}[1]{{#1}}
\providecommand{\urlprefix}{URL }
\expandafter\ifx\csname urlstyle\endcsname\relax
  \providecommand{\doi}[1]{DOI~\discretionary{}{}{}#1}\else
  \providecommand{\doi}{DOI~\discretionary{}{}{}\begingroup
  \urlstyle{rm}\Url}\fi

\bibitem{ashish2008situational}
Ashish, N., Eguchi, R., Hegde, R., Huyck, C., Kalashnikov, D., Mehrotra, S.,
  Smyth, P., Venkatasubramanian, N.: Situational awareness technologies for
  disaster response.
\newblock In: Terrorism Informatics, pp. 517--544. Springer (2008)

\bibitem{atzori2010internet}
Atzori, L., Iera, A., Morabito, G.: {The internet of things: A survey}.
\newblock Computer networks \textbf{54}(15), 2787--2805 (2010)

\bibitem{blandford2004situation}
Blandford, A., Wong, B.W.: Situation awareness in emergency medical dispatch.
\newblock International journal of human-computer studies \textbf{61}(4),
  421--452 (2004)

\bibitem{carroll2006awareness}
Carroll, J.M., Rosson, M.B., Convertino, G., Ganoe, C.H.: Awareness and
  teamwork in computer-supported collaborations.
\newblock Interacting with computers \textbf{18}(1), 21--46 (2006)

\bibitem{chase1973mind}
Chase, W.G., Simon, H.A.: The mind's eye in chess  (1973)

\bibitem{cowlard2010sensor}
Cowlard, A., Jahn, W., Abecassis-Empis, C., Rein, G., Torero, J.L.: {Sensor
  Assisted Fire Fighting}.
\newblock Fire Technology \textbf{46}(3), 719--741 (2010)

\bibitem{fliszkiewicz2014evaluation}
Fliszkiewicz, M., Krasuski, A., Krenski, K.: Evaluation of a heat release rate
  based on massively generated simulations and machine learning approach.
\newblock Proceeding of FedCSIS 2014 Conference, Warsaw  (2014)

\bibitem{gorman2006measuring}
Gorman, J.C., Cooke, N.J., Winner, J.L.: Measuring team situation awareness in
  decentralized command and control environments.
\newblock Ergonomics \textbf{49}(12-13), 1312--1325 (2006)

\bibitem{sff}
Grant, C., Hamins, A., Nelson, B., Koepke, G.: {Research Roadmap for Smart Fire
  Fighting. NIST Special Publication 1191}.
\newblock Tech. rep. (2015)

\bibitem{grorud2008national}
Grorud, L.J., Smith, D.: {The National Fire Fighter Near-Miss Reporting. Annual
  Report 2008}.
\newblock In: An exclusive supplement to FireRescue magazine, pp. 1--24.
  Elsevier Public Safety (2008)

\bibitem{jahn2012forecasting2}
Jahn, W., Rein, G., Torero, J.: Forecasting fire dynamics using inverse
  computational fluid dynamics and tangent linearisation.
\newblock Advances in Engineering Software \textbf{47}(1), 114--126 (2012)

\bibitem{kahneman2009conditions}
Kahneman, D., Klein, G.: Conditions for intuitive expertise: a failure to
  disagree.
\newblock American Psychologist \textbf{64}(6), 515 (2009)

\bibitem{keller1989role}
Keller, L.R.: The role of generalized utility theories in descriptive,
  prescriptive, and normative decision analysis.
\newblock Information and Decision Technologies \textbf{15}(4), 259--271 (1989)

\bibitem{khaitan2014design}
Khaitan, S.K., McCalley, J.D.: {Design techniques and applications of
  cyberphysical systems: A survey}  (2014)

\bibitem{klein2008naturalistic}
Klein, G.: Naturalistic decision making.
\newblock {Human Factors: The Journal of the Human Factors and Ergonomics
  Society} \textbf{50}(3), 456--460 (2008)

\bibitem{klein1999sources}
Klein, G.A.: Sources of power: How people make decisions.
\newblock MIT press (1999)

\bibitem{klein1986rapid}
Klein, G.A., Calderwood, R., Clinton-Cirocco, A.: Rapid decision making on the
  fire ground.
\newblock In: Proceedings of the Human Factors and Ergonomics Society annual
  meeting, vol.~30, pp. 576--580. SAGE Publications (1986)

\bibitem{korbel2013locfusion}
Korbel, P., Wawrzyniak, P., Grabowski, S., Krasinska, D.: Locfusion
  api-programming interface for accurate multi-source mobile terminal
  positioning.
\newblock In: FedCSIS, pp. 819--823 (2013)

\bibitem{krasuski2014framework}
Krasuski, A.: A framework for dynamic analytical risk management at the
  emergency scene. from tribal to top down in the risk management maturity
  model.
\newblock In: Federated Conference on Computer Science and Information Systems
  (FedCSIS), pp. 323--330. IEEE (2014)

\bibitem{krasuski2013sensory}
Krasuski, A., Jankowski, A., Skowron, A., Slezak, D.: From sensory data to
  decision making: A perspective on supporting a fire commander.
\newblock In: 2013 IEEE/WIC/ACM International Joint Conferences on Web
  Intelligence (WI) and Intelligent Agent Technologies (IAT), pp. 229--236.
  IEEE (2013)

\bibitem{krasuski2012method}
Krasuski, A., Kre{\'n}ski, K., {\L}azowy, S.: A method for estimating the
  efficiency of commanding in the state fire service of poland.
\newblock Fire Technology \textbf{48}(4), 795--805 (2012)

\bibitem{meehl149clinical}
Meehl, P.E.: Clinical vs. statistical prediction: a theoretical analysis and a
  review of the evidence.
\newblock Minneapolis, MN, US: University of Minnesota Press \textbf{149}
  (1954)

\bibitem{meina2015tagging}
Meina, M., Janusz, A., Rykaczewski, K., Slezak, D., Celmer, B., Krasuski, A.:
  Tagging firefighter activities at the emergency scene: Summary of aaia'15
  data mining competition at knowledgepit.
\newblock In: Federated Conference on Computer Science and Information Systems
  (FedCSIS), 2015, pp. 367--373. IEEE (2015)

\bibitem{meina2015}
Meina, M., Krasuski, A., Rykaczewski, K.: Model fusion for inertial-based
  personal dead reckoning systems.
\newblock In: Sensors Applications Symposium (SAS), 2015 IEEE, pp. 1--6 (2015).
\newblock \doi{10.1109/SAS.2015.7133658}

\bibitem{mendoncca2009group}
Mendon{\c{c}}a, D., Gu, Q., Osatuyi, J.: Group information foraging in
  emergency response.
\newblock In: Life Saver project workshop, Set{\'u}bal, Portugal (2009)

\bibitem{overholt2012characterizing}
Overholt, K.J., Ezekoye, O.A.: {Characterizing Heat Release Rates Using an
  Inverse Fire Modeling Technique}.
\newblock Fire Technology \textbf{48}(4), 893--909 (2012)

\bibitem{sorell2003ii}
Sorell, T.: {II--Morality and Emergency}.
\newblock In: Proceedings of the Aristotelian Society (Hardback), vol. 103, pp.
  21--37. Wiley Online Library (2003)

\bibitem{tversky1974judgment}
Tversky, A., Kahneman, D.: {Judgment under uncertainty: Heuristics and biases}.
\newblock Science \textbf{185}(4157), 1124--1131 (1974)

\end{thebibliography}

\end{document}